\documentclass[10pt,letterpaper]{article}
\usepackage[top=0.85in,left=2.75in,footskip=0.75in,marginparwidth=2in]{geometry}

\usepackage{amsmath,amssymb}

\usepackage{changepage, graphicx}

\usepackage[utf8x]{inputenc}

\usepackage{cite}

\usepackage{hyperref}

\usepackage[table]{xcolor}

\newcolumntype{+}{!{\vrule width 2pt}}

\usepackage{floatrow}
\usepackage{changepage}

\usepackage[aboveskip=1pt,labelfont=bf,labelsep=period,singlelinecheck=off]{caption}

\makeatletter
\renewcommand{\@biblabel}[1]{\quad#1.}
\makeatother

\date{\today}

\begin{document}

{\large
\textbf{\newline{Learning from Label Proportions in Brain-Computer Interfaces: Online Unsupervised Learning with Guarantees
}} 
}
\newline
\\
David Hübner\textsuperscript{1*},
Thibault Verhoeven\textsuperscript{3},
Konstantin Schmid\textsuperscript{1},
Klaus-Robert Müller\textsuperscript{2,4},
Michael Tangermann\textsuperscript{1*},
Pieter-Jan Kindermans\textsuperscript{2*}
\\

\parindent 0pt
\textbf{1} Brain State Decoding Lab, Cluster of Excellence BrainLinks-BrainTools, Department of Computer Science, Albert-Ludwigs-University, Freiburg, Germany
\\
\textbf{2} Machine Learning Group, Berlin Institute of Technology, Berlin, Germany
\\
\textbf{3} Electronics and Information Systems, Ghent University, Ghent, Belgium
\\
\textbf{4}
Department of Brain and Cognitive Engineering, Korea University, Seoul, Korea
\bigskip

*\textbf{Corresponding authors}: \\
david.huebner@blbt.uni-freiburg.de \\
michael.tangermann@blbt.uni-freiburg.de \\
p.kindermans@tu-berlin.de

\section*{Abstract}

\textbf{Objective: }Using traditional approaches, a Brain-Computer Interface (BCI) requires the collection of calibration data for new subjects prior to online use. Calibration time can be reduced or eliminated e.g.~by subject-to-subject transfer of a pre-trained classifier or unsupervised adaptive classification methods which learn from scratch and adapt over time. While such heuristics work well in practice, none of them can provide theoretical guarantees. Our objective is to modify an event-related potential (ERP) paradigm to work in unison with the machine learning decoder, and thus to achieve a reliable unsupervised calibration-less decoding with a guarantee to recover the true class means.

\textbf{Method: }
We introduce learning from label proportions (LLP) to the BCI community as a new unsupervised, and easy-to-implement classification approach for ERP-based BCIs. The LLP estimates the mean target and non-target responses based on known proportions of these two classes in different groups of the data. We modified a visual ERP speller to meet the requirements of the LLP. For evaluation, we ran simulations on artificially created data sets and conducted an online BCI study with $N=13$ subjects performing a copy-spelling task.

\textbf{Results: }
Theoretical considerations show that LLP is guaranteed to minimize the loss function similarly to a corresponding supervised classifier. It performed well in simulations and in the online application, where 84.5\,\% of characters were spelled correctly on average without prior calibration.

\textbf{Significance: }
The continuously adapting LLP classifier is the first unsupervised decoder for ERP BCIs guaranteed to find the true class means. This makes it an ideal solution not only to avoid a tedious calibration, but also to tackle non-stationarities in the data. Additionally, LLP works on complementary principles compared to existing unsupervised methods, opening the door for their further enhancement when combined with LLP.
\parindent 10pt
\section{Introduction}
A brain-computer interface (BCI) is a neurotechnological solution to control a software or a physical device, e.g.~allowing physically challenged users to send messages to caregivers or to operate a robotic device without muscular input. In this work, we focused on BCI applications based on event-related potentials (ERPs) measured by electroencephalography (EEG). ERPs are evoked transient brain patterns observed after, but not limited to, external stimulation events. In the field of BCI, the visual highlighting of symbols on a computer screen is the most often used stimulus modality~\cite{farwell1988talking}, but also non-visual stimuli like sounds~\cite{schreuder2010new, furdea2009auditory, simon2015auditory}, or haptic stimuli~\cite{schreuder2012exploring, brouwer2010tactile} are suited for BCI control. By assigning control commands to symbols on a screen, the user can execute a command by focusing attention onto the highlighting events corresponding to the desired symbol. Visual ERP paradigms have been used for different applications, e.g.~for spelling ~\cite{farwell1988talking, blankertz2010bbcitoolbox}, web browsing~\cite{bensch2007nessi}, games~\cite{finke2009mindgame, blankertz2010bbcitoolbox, congedo2011brain, kaplan2013adapting}, browse and share pictures~\cite{tangermann2011optimized}, predicting emergency brakes in a driving scenario~\cite{blankertz2016berlin} and artistic expression through painting~\cite{munssinger2010brain}. ERP-based BCIs have several desirable features~\cite{fazel2012p300}: they are relatively fast, effective for most healthy users~\cite{guger2009many} and usable for patients~\cite{munssinger2010brain, sellers2006p300}. The visual interfaces are easy to grasp and require virtually no subject training. Consequently, ERP-based BCIs are the most widely used BCI paradigms.

The machine learning decoder in ERP paradigms has to discriminate single stimulus events between attended (\textit{target} events), or non attended (\textit{non-target} event). Three reasons make this a very challenging task~\cite{wolpaw2002brain}. First, ERP features of the EEG are obscured by a high noise level and low signal amplitude, resulting in a bad signal-to-noise ratio (SNR). Second, non-stationarities can occur in the data. These are caused by varying factors~\cite{shenoy2006towards}: by motivation, level of attention, fatigue, mental state, learning, changes in contact impedances of EEG electrodes and others. Third, the statistical properties of the ERP signals and the background EEG differ from subject to subject. Thus, subject-specific data is necessary to obtain optimal decoding performance. Unsupervised methods, in contrast to supervised methods, have the benefit of skipping the calibration session to collect labelled data \cite{kindermans2012bayesian, kindermans2014true, kindermans2014integrating} and learn directly on unlabelled data collected during the online use of the decoder \cite{vidaurre2011toward}. This is desirable because it reduces the preparation time, avoids the problem of potentially measuring wrongly labelled data in the calibration session, e.g.~due to mistakes in the communication between instructor and user, and circumvents the difficulty of dealing with changes in the distribution of class-informative ERP features in the data from the calibration phase to the online application. These may occur due to changes in the human-computer interaction ~\cite{vidaurre2011toward}, e.g.~by introducing performance feedback to the user.

However, the ability to learn from unlabelled examples comes at a price. Until now, there is no unsupervised algorithm which is guaranteed to converge to the right solution even if sufficient data is available. Several attempts have previously been made to reduce the calibration time. Transfer learning approaches combined with unsupervised adaptation methods are able to drastically decrease the calibration time in ERP paradigms \cite{kindermans2014integrating, barachant2014plug, lu2009unsupervised, jayaram2016transfer, vidaurre2011toward, kindermans2014true}. Subject independent methods have been successfully introduced where a new experiment of the same subject~\cite{krauledat2008towards} or of a new subject~\cite{FazDahSamBieMul15, FazPopDanBlaMueGro09} can profit from the data base of existing subjects. While these methods generally work well in practice, they partly rely on random initialisations~\cite{kindermans2014integrating, kindermans2014true} generating an additional source of uncertainty.

The main contribution of this paper is the introduction of learning from label proportions (LLP) to the field of BCI. LLP is a recently proposed unsupervised classification method~\cite{quadrianto2009estimating}. It is capable of learning from unlabelled data and is guaranteed to recover the same target and non-target means as in a scenario where label information would be present. To the best of our knowledge, this is the first unsupervised algorithm for classification of ERP signals which, under the assumption of independent and identically distributed (IID) data points, is guaranteed to minimize several loss functions. 

To grasp the main idea of LLP, consider the following scenario. We want to estimate the average weight of men and that of women. We are not able to weigh people individually but are given aggregated data from groups of the population. There are two groups of people. The first one consists of 50 men and 40 women and has a total weight of 6600\,kg, the second one comprises 40 men and 60 women and has a total weight of 7100\,kg. The avid reader will quickly realize that the average weight can easily be computed by solving a linear system of 2 equations yielding a men's average weight of 80\,kg and a women's average weight of 65\,kg. Surprisingly, the mean weight for men and women can be computed without actually knowing the weight of a single individual man or woman. This yields an unsupervised method where label information is not required. It is sufficient to know the group-wise means and the proportional presence of each class --- man and woman --- in the different groups. In a similar fashion to the example above, LLP can be applied to an ERP paradigm to reconstruct the mean target and non-target ERP responses which can then be used to train a classifier and classify individual stimuli. 

In the following sections, the LLP algorithm will be discussed. First, the theoretical background and properties will be derived in the methods section. Secondly, computer simulations will show that the approach works well on different artificial BCI data sets. However, as certain assumptions cannot be tested with BCI simulations, the outcome of an online study with $N = 13$ subjects will be presented as a proof of principle. Additionally, the ramp-up performance will be compared with the EM-algorithm by Kindermans et al.~\cite{kindermans2014true}, and possible improvements and application scenarios will be discussed.

\section{Materials and Methods}
\subsection{Learning from Label Proportions }\label{sec:llp_theory}
\textbf{Theoretical Motivation: }
In supervised binary classification, the goal is to discriminate two classes. In machine learning, we optimize a loss function depending on the data $\mathbf{x}_i$ and the labels $y_i \in \{−1, 1\}$ where $i=1 \ldots N$ denotes the different samples. In the following, we will assume that a linear classifier $f(\mathbf{x}) = \mathbf{w}^T \mathbf{x}$ is used which assigns a sample $\mathbf{x}_j$ to class 1 if $f(\mathbf{x}_j) \geq 0$ and to class 2 if $f(\mathbf{x}_j)<0$. Because the classification loss cannot be optimized directly since it is discrete and non-convex, machine learning methods optimise a surrogate loss function instead. For a specific subset of losses called \textit{symmetric proper scoring losses} which include the logistic loss, the 0-1 loss and the square loss, we can rewrite the loss function in a form that depends only on the class means and the input data~\cite{patrini2014no_label_no_cry}. Below, we have rewritten the square loss as an explicit example.
\begin{eqnarray*}
	\sum_{i=1}^{N}{(\mathbf{w}^{T} \mathbf{x}_i - y_i)}^2  
	= \sum_{i=1}^{N}  	\left( {(\mathbf{w}^{T} \mathbf{x}_i)}^2 + 1 \right)
	- 2 \mathbf{w}^{T} \left( \sum_{i_+} \mathbf{x}_{i_+} - \sum_{i_-} \mathbf{x}_{i_-} \right)
\end{eqnarray*}
Here $i_+$ denotes samples from the class with label $+1$ and $i_-$ denotes samples from the class with label $-1$. It is clear that the term $\sum_{i=1}^{N} ({(\mathbf{w}^{T} \mathbf{x}_i)}^2 + 1)$ does not depend on label information. The second term can now be reformulated as
\begin{eqnarray*}
2 \mathbf{w}^{T} \left( \sum_{i_+} \mathbf{x}_{i_+} - \sum_{i_-} \mathbf{x}_{i_-} \right)
= 2 \mathbf{w}^{T} ( N_+ \hat{\boldsymbol{\mu}}_+ - N_- \hat{\boldsymbol{\mu}}_- )
\end{eqnarray*}
where $\hat{\boldsymbol{\mu}}_+$ and $\hat{\boldsymbol{\mu}}_-$ indicate the average feature vector of the positive class and negative class, respectively. $N_+$ and $N_-$ represent the number of samples in each class. From this equation, it becomes clear that the optimization problem can be solved by merely \textit{knowing the class means and number of samples per class} without explicit label information. In the following sections, we will explain how the mean map algorithm~\cite{quadrianto2009estimating} for learning from label proportions can be used to reconstruct these means.

\textbf{Main Concept: }
Consider a two-class problem and G groups of data where each group is a mixture of these two classes with known mixture ratios contained in $\boldsymbol \Pi$. The expected values of the feature vectors in the groups $\boldsymbol{\mu}_1, \boldsymbol{\mu}_2, \ldots, \boldsymbol{\mu}_G$  can then be expressed as a function of the class means $\boldsymbol{\mu}_+, \boldsymbol{\mu}_-$ as follows:
\begin{eqnarray*}
\begin{bmatrix} \boldsymbol{\mu}_1  \\ \vdots \\ \boldsymbol{\mu}_G\end{bmatrix} 
=\boldsymbol \Pi 
\begin{bmatrix} \boldsymbol{\mu}_+  \\ \boldsymbol{\mu}_-\end{bmatrix}, \quad \boldsymbol \Pi = \begin{bmatrix} {\pi_+^1} & {\pi_-^1}  \\  \vdots &  \vdots \\ 	{\pi_+^G} & {\pi_-^G}	\end{bmatrix}
\end{eqnarray*}
To obtain an empirical estimate of the group means $\boldsymbol{\mu}_1, \boldsymbol{\mu}_2, \ldots, \boldsymbol{\mu}_G$, we do not need label information. These quantities can then be used to approximate the class means by using the psuedoinverse of $\boldsymbol\Pi$.
\[
\begin{bmatrix} \boldsymbol{\mu}_+  \\ \boldsymbol{\mu}_-\end{bmatrix}=(\boldsymbol\Pi^T\boldsymbol\Pi)^{-1}\boldsymbol\Pi^T\begin{bmatrix} \boldsymbol{\mu}_1  \\ \vdots \\ \boldsymbol{\mu}_G\end{bmatrix} 
\] Hence, by solving the resulting system of linear equations, we can get an estimation $\tilde{\boldsymbol{\mu}}_+, \tilde{\boldsymbol{\mu}}_-$ of the true class means $\boldsymbol{\mu}_+, \boldsymbol{\mu}_-$. In the BCI application, these two classes are target and non-target. The implicit \textit{homogeneity assumption} in this formulation is that $\boldsymbol{\mu}_+, \boldsymbol{\mu}_-$ are the same for each group, i.e.~the feature distributions for both, target and non-target samples, are independent of the group. 

\textbf{Guaranteed Convergence: }
If we assume that each feature $x_1,x_2,\ldots,x_N$ is drawn independently from an identical distribution (IID) with finite expected value $\mu$ and variance $\sigma$, then the central limit theorem (CLT) states that the sample average $S_N$ is Gaussian distributed for large N.
 
\begin{eqnarray*}
S_N := \frac{x_1 + \ldots + x_N}{N} \sim \mathcal{N}(\mu, \frac{\sigma^2}{N}) 
\end{eqnarray*}

This implies that, given enough data, the approximations $\tilde{\boldsymbol{\mu}}_1, \ldots, \tilde{\boldsymbol{\mu}}_G$ converge to $\boldsymbol{\mu}_1, \ldots, \boldsymbol{\mu}_G$ in our scenario. After solving the linear system, we therefore have an estimation of the class-wise means which is \textit{guaranteed to converge} for $N \rightarrow \infty$. Hence, we have an unsupervised classifier that, under the assumption of IID and homogeneity, is guaranteed to minimize the squared error loss. Additionally, there also exists a version of the mean-map algorithm using a manifold regulariser that performs better than this version under a violation of the homogeneity assumption~\cite{patrini2014no_label_no_cry}.

\textbf{Noise Amplification Factor: }
Like other unsupervised algorithms, LLP performs worse compared to a supervised classifier when only a limited number of data points is available. In the LLP case, we can directly quantify the difference which depends on the number of groups ($G$) and the inverse of the mixture matrix $\boldsymbol \Pi^{-1}$. Note that we use the pseudo-inverse if $G > 2$ and let it be given by:

\begin{eqnarray*}
\boldsymbol \Pi^{-1}  = \begin{bmatrix} {\nu_+^1} & \ldots &  {\nu_+^G} \\ {\nu_-^2} & \ldots & {\nu_-^G} 	\end{bmatrix}
\end{eqnarray*}
Now using the properties of the variance, the variance of  $\boldsymbol{\mu}_{+}$ can be computed as:
\begin{eqnarray*}
\operatorname{Var} (\boldsymbol{\mu}_{+}) 
= \operatorname{Var} \left(\Pi^{-1} \begin{bmatrix} \boldsymbol{\mu}_1  \\ \vdots \\ \boldsymbol{\mu}_G\end{bmatrix}\right) 
= \operatorname{Var} \left( \sum_{k=1}^G \nu_{+}^k \boldsymbol{\mu}_k \right) 
=  \sum_{k=1}^G {(\nu_{+}^k)}^2 \operatorname{Var} (\boldsymbol{\mu}_k) 
\end{eqnarray*}
and analogously for $\boldsymbol{\mu}_{-}$. This implies that the variance of each estimator is amplified by the square of the pseudoinverse coefficients. The variance of each feature  ${\mu}_{k}^j$ of the group means $\boldsymbol{\mu}_k$ is given by
\begin{eqnarray*}
\operatorname{Var} ({\mu}_{k}^j)
= \frac{\sigma^2}{\frac{N}{G}}, \quad k=1 \ldots G
\end{eqnarray*}
where $\frac{N}{G}$ is the fraction of data available per group. To quantify the increased variance of the class-wise mean estimation compared to the original variance of $\frac{\sigma^2}{N}$ , we define the \textit{noise amplification factor} (NAF) as the number of groups G multiplied with the squared Frobenius norm which is the sum of all squared coefficients in $\boldsymbol \Pi^{-1}$.
\begin{eqnarray*}
\operatorname{NAF} := G \cdot \sum_{c \in \{+,-\}} \ \sum_{k=1}^G {(\nu_{c}^k)}^2
\end{eqnarray*}
In the result section, we will evaluate the performance for different numbers of groups and mixture matrices.

\subsection{Sequence Generation and Spelling Interface}
Now, we address how these different groups of data can be obtained. We propose to tune the stimulus presentation to maximize the power of the machine learning algorithm. To achieve this symbiosis, we started from the original visual P300 speller~\cite{farwell1988talking} and did several modifications. First, an additional column was added to increase the total number of symbols resulting in a 6 x 7 grid. We included all letters of the alphabet plus the symbols "\textvisiblespace" "." "," "!" "?" "$\leftarrow$" and 10 "\#" symbols which account for visual blanks. The meaning of those is explained further below.

To generate two different groups of data ($G=2$) in the visual speller, we used the stimulus presentation paradigm created by Verhoeven et al.~\cite{Verhoeven2015Towards}. This paradigm is flexible in the sense that it can generate sequences with a desired mixture ratio of target and non-target stimuli. At the same time, it uses a heuristic to increase the signal-to-noise ratio in the stimulus responses by avoiding the two most common spelling errors: adjacency distraction and double flashes. Our modification requires two distinct sequences with differing target to non-target ratios --- as these ratios form the label proportions exploited by LLP. Sequences of 8 stimuli containing 3 targets and sequences of 18 stimuli containing 2 targets were generated. The resulting linear system is

\begin{eqnarray*}
\left\{
\begin{aligned}
\boldsymbol{\mu}_1 &= \phantom{1} \frac{3}{8} \boldsymbol{\mu}_+  + \frac{5}{8} \boldsymbol{\mu}_- \\
\boldsymbol{\mu}_2 &= \frac{2}{18} \boldsymbol{\mu}_+ + \frac{16}{18} \boldsymbol{\mu}_-
\end{aligned}
\right.
\end{eqnarray*}

where $\boldsymbol{\mu}_1$ and $\boldsymbol{\mu}_2$ are the sequence-wise averages. For this simple configuration, the mean target $\boldsymbol{\mu}_+$ and non-target ERP responses $\boldsymbol{\mu}_-$ can directly be computed by solving the linear system yielding the following two equations. 
\begin{eqnarray*}
\left\{
\begin{aligned}
\boldsymbol{\mu}_+ &= \phantom{-}3.37 \, \boldsymbol{\mu}_1 -2.37 \, \boldsymbol{\mu}_2 \\
\boldsymbol{\mu}_- &=-0.42 \, \boldsymbol{\mu}_1 +1.42 \, \boldsymbol{\mu}_2
\end{aligned}
\right.
\end{eqnarray*}  
  
In ERP terminology, a trial corresponds to the selection of a single command. In our approach a trial consisted of 4 sequences of length 8 and 2 sequences of length 18, totalling to 68 highlighting events. The 16 targets and 52 non-targets highlighted per trial each resulted in an ERP response in the EEG, leading to 68 ERP \textit{epochs}.

A few additional measures were taken to comply with our assumption, that ERP responses are distributed identically and homogeneously per group. First of all, it is known that the response upon a stimulus event is influenced by its brightness and thus by the number of symbols highlighted within that stimulus event~\cite{johannes1995luminance}. To equalize the number of symbols among stimuli, the 10 "\#" symbols were introduced in addition to the standard symbols. Adding them balances the brightness of those stimuli containing less symbols otherwise. As they never convey information, they take the role of non-target symbols, and do not alter the mixture ratio of the sequences. Using the enlarged symbol set, each event highlighted 12 out of 42 symbols. 
The second precaution taken is that sequences from both groups were randomly interleaved on a trial level. This avoids the violation of the homogeneity assumption, e.g.~non-stationarity in the feature distribution within one trial or a modulation of the P300 amplitude because of differences in the target-to-target interval~\cite{gonsalvez2002p300}. 

For stimulus presentation, a salient highlighting method proposed by Tangermann et al.~\cite{tangermann2011optimized} was implemented. It uses a combination of brightness enhancement, rotation, enlargement and a trichromatic grid overlay. An example of the highlighting scheme and the spelling matrix is shown in Fig~\ref{Fig:exp_structure}. 

\subsection{Experimental Protocol, Data Quantity and Task Timing}
\begin{figure}[H]
\centering
    \includegraphics[width=\textwidth]{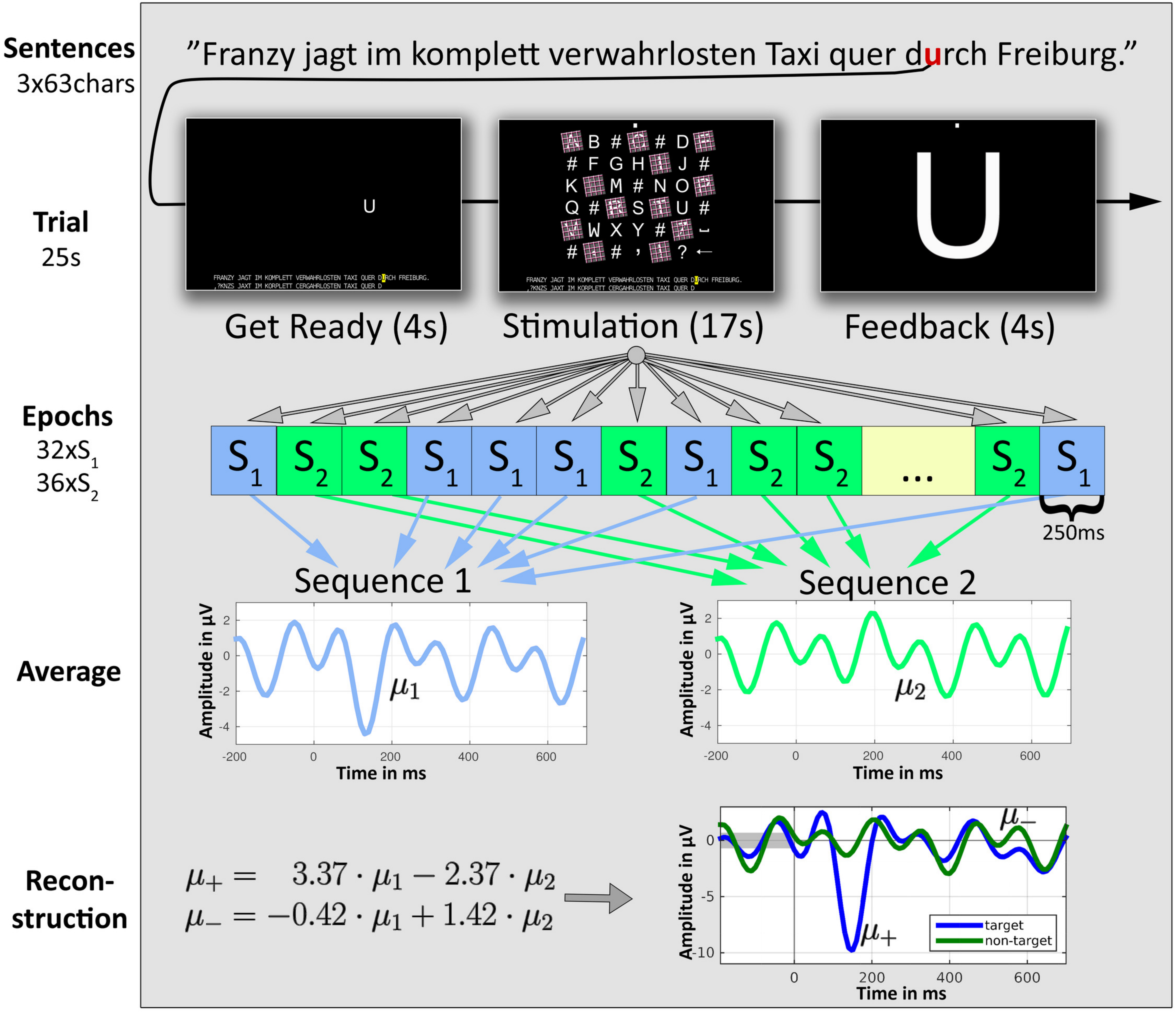}
    \caption{\textbf{Scheme of the experimental structure and LLP classifier.} \color{darkgray} \\ Top to bottom: The sentence "Franzy jagt ..." is spelled three times. To spell a single character in one trial, 68 highlighting events occur, with 32 belonging to sequence 1 and 36 belonging to sequence 2. The resulting 68 ERP responses (epochs) are averaged for each sequence, and these averages are exploited to reconstruct the mean target and non-target ERP responses.}
    \label{Fig:exp_structure}
\end{figure}

The subjects were asked to spell the sentence: "\textsc{Franzy jagt im komplett verwahrlosten Taxi quer durch Freiburg}". Each subject spelled the sentence three times. The stimulus onset asynchrony (SOA) was 250\,ms (corresponding to 15 frames on the LCD screen utilized) while the stimulus duration was 100\,ms (corresponding to 6 frames on the LCD screen utilized). For each character, 68 highlighting events occurred and a total of 63 characters were spelled three times. This resulted in a total of $68 \cdot 63 \cdot 3 = 12852$ EEG epochs per subject. Spelling one character took around 25\,s including 4\,s for cueing the current symbol, 17\,s for highlighting and 4\,s to provide feedback to the user. Assuming a perfect decoding, these timing constraints would allow for a maximum spelling speed of 2.4 characters per minute. Fig~\ref{Fig:exp_structure} shows the complete experimental structure and how LLP is used to reconstruct average target and non-target ERP responses.

\subsection{EEG Data Acquisition}
Subjects were placed in a chair at 80\,cm distance from a 24-inch flat screen. EEG signals from 31 passive Ag/AgCl electrodes (EasyCap) were recorded, which were placed approximately equidistantly according to the extended 10–20 system, and whose impedances were kept below 20\,k$\Omega$. All channels were referenced against the nose. The signals were registered by multichannel EEG amplifiers (BrainAmp DC, Brain Products) at a sampling rate of 1~kHz. To control for vertical ocular movements and eye blinks, we recorded with an EOG electrode placed below the right eye and referenced against the EEG channel Fp2 above the eye. In addition, pulse and breathing activity were recorded. However, the EOG, pulse and breathing signals did not enter the further analysis. Markers obtained from an optical sensor on the screen indicated the exact starting time point of each highlighting event.

\noindent \textbf{The data is freely available online at} \newline \url{http://doi.org/10.5281/zenodo.192684}.

\subsection{Preprocessing, Classification and Scoring} \label{preprocessing}
\textbf{Preprocessing: }
To process the data in the online experiment and during offline re-analysis, the BBCI Toolbox was used~\cite{blankertz2010bbcitoolbox}. In both cases, the collected data was bandpass filtered with a third order Chebyshev Type II filter between 0.5 and 8\,Hz and downsampled to 100\,Hz. Epochs were windowed to [-200, 700]\,ms relative to the stimulus onset and corrected for baseline shifts observed in the interval [-200, 0]\,ms. After dismissing channels Fp1 and Fp2, features describing the elicited transient potentials were extracted from the remaining 29 EEG channels. Per channel, the mean amplitudes of six intervals ([50, 120], [121, 200], [201, 280], [281, 380], [381, 530] and [531, 700]\,ms) were computed, resulting in a representation of each epoch by $6 \cdot 29 = 174$ features.

\textbf{Classification: }
Based on the features collected up to the current point of each sentence, the LLP algorithm was applied online to estimate the class means $\boldsymbol{\mu}_{+}$ and $\boldsymbol{\mu}_{-}$. Additionally, the pooled (global) covariance matrix $\boldsymbol \Sigma$ on combined data of both classes was estimated using shrinkage-regularization as initially proposed by Ledoit\&Wolf~\cite{ledoit2004shrinkage} and first applied in BCI by Vidaurre et al.~\cite{vidaurre2009time}, see also~\cite{blankertz2011single}. The shrinkage parameter was chosen automatically using the Ledoit--Wolf formula~\cite{schafer2005shrinkage}. Based on the reconstructed class means and the pooled covariance matrix, the projection vector $\mathbf w$ was computed as

\begin{equation*}
\mathbf w = \boldsymbol{\Sigma}^{-1} \left( \boldsymbol{\mu}_{+} - \boldsymbol{\mu}_{-} \right)
\end{equation*}

and applied to the features of a new epoch $\mathbf{x_{new}}$ as $f(\mathbf{x_{new}})=\mathbf{ w^T \mathbf{x_{new}}}$. Technically, this can be understood as a least square classifier with re-scaled outputs. To select a symbol in each trial, classifier outputs were summed up for each symbol and the symbol with the highest sum was chosen. Note that this decision does not depend on the bias term, because the same bias is summed up for each symbol and thus, its effect cancels out when taking the maximum. Visual blanks were excluded from being chosen as selected symbols. 

The classifier was reset and started from scratch for each of the three spellings of the sentence "\textsc{Franzy jagt ...}" in the online experiment. After collecting the data of a new character, the classifier was retrained. Label information (target / non-target role of characters) were used exclusively to evaluate the performance during offline analyses, but never to train the LLP classifier during online use.

In the supervised scenario, which was used solely in the offline analysis for comparison, class-wise means and the class-wise covariance matrices were calculated based on label information. Shrinkage-regularization and the projection vector were computed as described before. We refer to this classifier as shrinkage-LDA~\cite{blankertz2011single}.

\textbf{Scoring: }
To assess the performance of any classifier, the area under the curve (AUC) of the receiver-operator characteristics curve (ROC) for classifying target vs.~non-target epochs was calculated. The AUC values can range between 0 and 1, with a theoretical chance level of 0.5. An AUC value of 1 indicates perfect separation between the two classes. The AUC can be seen as the probability that the output of the LDA ranks a target higher than a non-target. We chose AUC as it is non-parametric and independent of a classifier bias. When we report an AUC, it always considers the binary target vs. non-target classification task. In addition to the AUC metric, the percentage of correctly classified symbols is reported where appropriate to describe the performance in the online spelling application. The information transfer rate (ITR) was not considered as a performance metric, since the assumptions required to use ITR reliably are not met by ERP paradigms~\cite{thompson2014performance}. 

\subsection{Unsupervised Post Hoc Classification}\label{sec:posthoc}
An interesting feature of adaptive classifiers is that their quality improves over time as more and more unlabelled data becomes available during their online application. Hence, re-analysing previous trials may result in more accurate decoding results compared to the results obtained online. This so-called post hoc re-analysis can easily be included in an online experiment as done before by Kindermans et al.~\cite{kindermans2014true}. In applications like text spelling, the constant post hoc re-analysis may prove extremely beneficial to correct early spelling mistakes, which have to be accounted to the limited data at the start of the unsupervised experiment. In a real-life spelling task, the user would need to accept that early characters initially are misspelled by the system, but would probably be corrected at a later time point. Thus, for the user, it is a fruitful strategy to continue spelling the sentence despite of potential incorrectly decoded characters. 
 
\subsection{Subjects and Ethics}
Overall, 13 subjects (5 female, average age: 26 years, std: 1.5 years) were recruited. Only one subject (S2) had prior EEG experience. The EEG study was approved by the Ethics Committee of the University Medical Center Freiburg. Following the principles of the Declaration of Helsinki, written informed consent was obtained from the subjects prior to participation. One session took about 3 hours (including EEG set-up and washing the hair), and participants were compensated with 8 Euros per hour.

\subsection{Artifical Data Sets for Simulations}\label{sec:artificalDatasets}
In addition to data collected during an online experiment, we created artificial data sets for simulations. They were based on EEG data of two real ERP-BCI data sets. The first data set stemmed from a visual attention task with 6 possible choices. The second data set had been recorded with an auditory ERP paradigm with spatial cues similar to the AMUSE paradigm~\cite{schreuder2010new}. Both data sets had been recorded under an SOA of 250\,ms. In both paradigms, 4860 epochs were recorded with the same 5 young healthy volunteers each. These data sets were chosen for simulations in order to cover different SNR values --- the first data set has a very high SNR due to the small number of selectable items, while the auditory data set displays a low SNR compared to data obtained from visual ERP paradigms. For each data set, artificial sequences were created by assigning target and non-target epochs randomly to each of the new sequences based on a pre-defined mixture matrix $\boldsymbol{\Pi}$. This was done for different mixture matrices and a varying number of epochs ranging from 500 to 4860 where epochs were taken chronologically starting from the beginning of the experiment. For each of the mixing matrices and number of epochs, an LLP classifier based on the reconstructed means and the pooled covariance matrix was trained and tested in a 5-fold chronological cross-validation, see~\cite{blankertz2011single}. In contrast to our LLP online study, 64 channels were used, eye-artefacts were regressed out using the EOG channel \cite{parra2005recipes} and the intervals \mbox{[100 180]}, \mbox{[181 300]}, \mbox{[301 400]}, \mbox{[401 600]}, \mbox{[601 850]} and \mbox{[851 1200]}\,ms were used in case of the auditory study. The other pre-processing steps are the same as mentioned above.
 
\subsection{Bootstrapping}
A leave-one-out bootstrapping test was performed offline to assess whether the homogeneity assumption holds for the data recorded in the online experiment. The idea is to compute the similarity of a sample from sequence 1 to the average ERP response from sequence 1 and to the average ERP response from sequence 2. The similarity values allow to test whether the null hypothesis holds i.e.~that target and non-target responses follow the same distribution for both sequences. After applying the same preprocessing steps as mentioned before, we iterated over each target (non-target) epoch of sequence 1. The average target (non-target) ERP responses for both sequences were computed where the selected epoch was excluded when calculating the average of sequence 1. In the next step, the squared distance ($L_2$ - norm) between the selected epoch and the previously computed averages was calculated in the interval [0,700]~ms using all channels. A two-sided T-test was finally conducted to check, whether these distances differ significantly. This procedure was done separately for each class and subject, yielding a total of $2 \cdot 13 = 26$ tests.

\section{Results}
The result section is divided into three parts. First, we present the simulation results, subsequently the online study is evaluated and finally a comparison between LLP and the EM-algorithm by Kindermans et al.~\cite{kindermans2012bayesian, kindermans2014integrating, kindermans2014true} is presented. 

\subsection{Simulations}
\begin{figure}[H]

\begin{center}
\includegraphics[trim=0 0 160 0,clip,width=\textwidth]{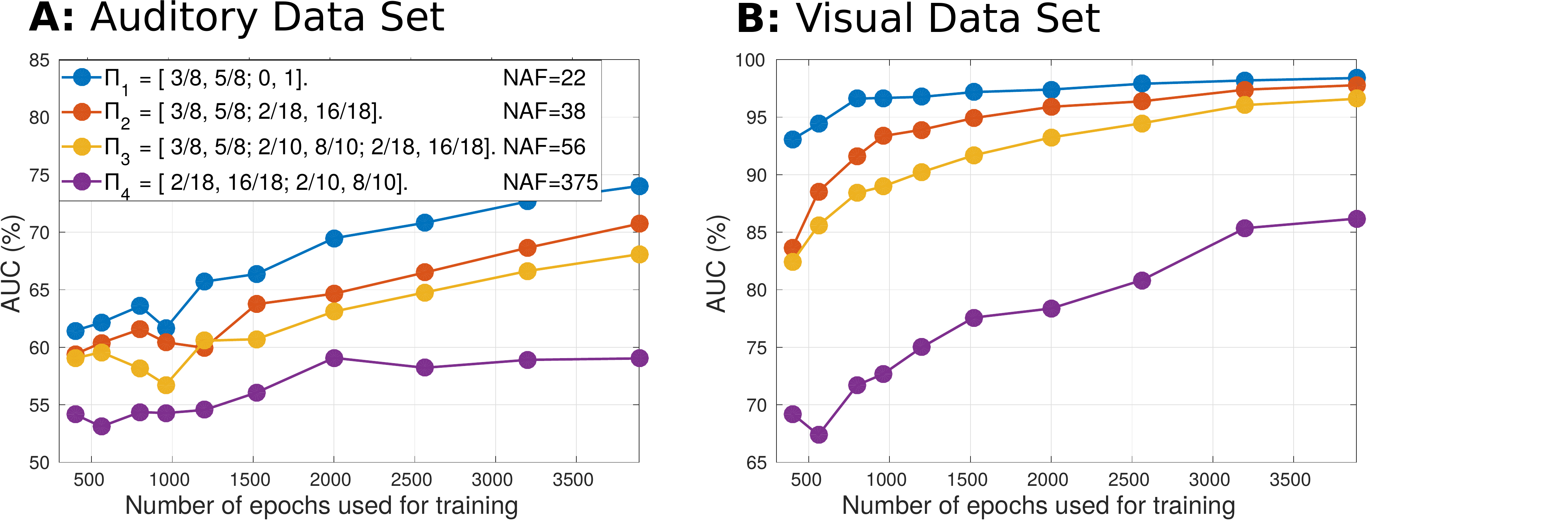}
\end{center}

\caption{\textbf{Classification results of LLP applied on artificial data sets generated from an auditory (A) and a visual (B) ERP paradigm.} \\ \color{darkgray}
For each artificial data set, the target vs non-target accuracy result for different mixing matrices $\boldsymbol{\Pi}_1$-$\boldsymbol{\Pi}_4$ is shown. In the notation of the mixing matrices, individual sequences are separated by commas. Per sequence, the first entry indicates the target ratio whereas the second one indicates the non-target ratio. NAF = Noise Amplification Factor.}
\label{Fig:sim_results}
\end{figure}

To evaluate the feasibility of LLP for BCI and to validate the theoretical considerations, we performed simulations on artificial data sets generated as described before. Creating artificial data sets allowed us to assess the effect of the mixing matrix on the quality of the class-wise mean reconstruction. Four different mixing matrices were used, ranging from one with extremely different target and non-target mixture ratios ($\boldsymbol{\Pi}_1$) to one with relatively similar sequences ($\boldsymbol{\Pi}_4$). The following observations can be made from the simulation results in Fig~\ref{Fig:sim_results}. First, the classifier performs well above chance level (50\%) on the auditory and visual data sets, indicating LLPs feasibility to reconstruct the class means. As expected, classification accuracy is much higher on the visual data (Fig~\ref{Fig:sim_results}\textbf{B}) where the algorithm reaches almost perfect performance for the well-conditioned mixing matrices $\boldsymbol{\Pi}_1$-$\boldsymbol{\Pi}_3$. The performance on the auditory data set is worse (Fig~\ref{Fig:sim_results}\textbf{A}), but still exceeds an AUC of 70\,\% for matrices $\boldsymbol{\Pi}_1$ and $\boldsymbol{\Pi}_2$ when provided with a sufficient amount of data. Second, we observe, that the performance over the four matrices can be ranked in the order of the ascending noise amplification factors, indicating that the NAF is a good parameter to characterize how well a mixing matrix determines the LLP performance. Finally, we observed that mixing matrix $\boldsymbol{\Pi}_3$ with three different sequences performs worse than some of the mixing matrices with two sequences. A reduction to only two out of three sequences, namely the one with the highest target ratio and the one with the highest non-target ratio, seems to be preferable over maintaining all three sequences. Dropping sequence [2/10, 8/10] from $\boldsymbol{\Pi}_3$ for instance yields $\boldsymbol{\Pi}_2$, which has a lower NAF and a higher performance. 

\vspace{-2.5cm}
The results obtained from the offline simulations indicate the feasibility to use LLP on data of ERP-BCIs. However, the central homogeneity assumption, i.e.~that target and non-target ERP responses follow the same distributions for all sequences, could not be tested in simulations. Hence, there is a need for an online study which we conducted with $N=13$ subjects.

\subsection{Online Experiment}
\subsubsection{Basic Neurophysiology and Supervised Performance} 
\begin{figure}[H]
\begin{center} 
\includegraphics[width=1\textwidth]{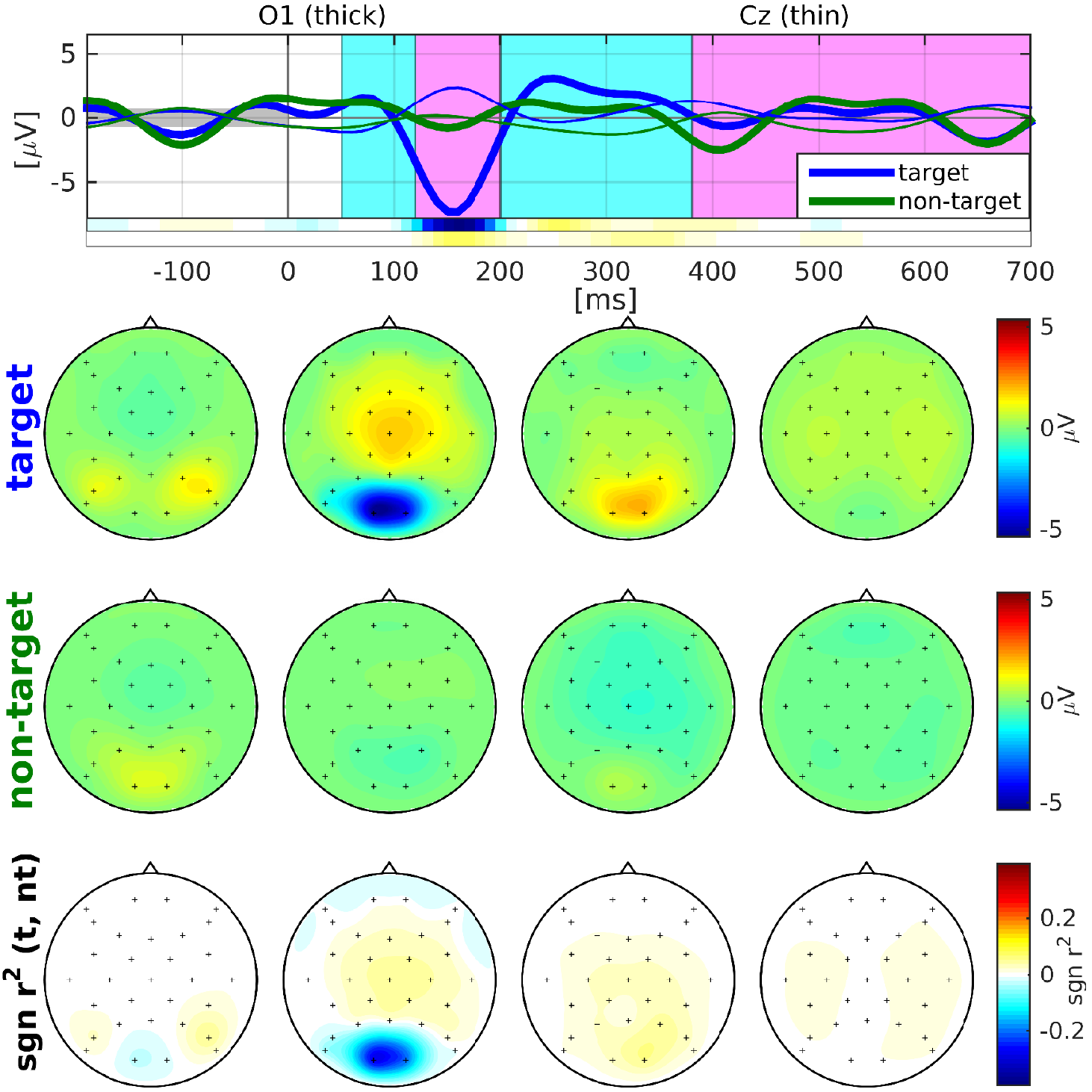} 
\end{center}
\caption{\textbf{Grand average (N=13) visual ERP response.}\\
\color{darkgray} Top row: Average responses evoked by visual target (blue) and non-target (green) stimuli in the occipital channel O1 (thick) and the central channel Cz (thin). Prior to averaging, a baseline correction was performed based on data within the interval [-200,0]\,ms. The signed $R^2$ values for channels O1 and Cz over time are provided by two horizontal color bars. Their scale is identical to the scale of plots in the last row. Middle rows: Scalp plots visualising the spatial distribution of mean target and non-target responses within four selected time intervals: [50 120], \mbox{[120 200]}, \mbox{[201 380]} and [381 700]\,ms relative to stimulus onset. Bottom row: Scalp plots with signed $R^2$ values indicate spatial areas with high class-discriminative information.}
\label{Fig:grand_average}
\end{figure}

First, we inspected the class-wise visual ERP responses to assess the quality of the data of the online study. They are provided as grand average responses in Fig.~\ref{Fig:grand_average}. The rhythmic characteristic of the non-target responses generally reflects the SOA of 250\,ms. We found a strong early negative ERP  upon target stimuli over the occipital lobe (hereafter called N150) at around 150\,ms for almost all subjects with an average amplitude of around $-8\,µV$. For non-target stimuli, the N150 was very reduced. The late positivity of targets (hereafter called P300) in the central electrodes is rather late and weak with an average peak time around 400\,ms and an average amplitude of only around $2\,µV$. Table~\ref{tab:neurophysiology} lists the amplitudes and peak latencies per subject observed for channels O1 (for the N150) and Cz (for the P300). 

\begin{table}[H]
\caption{\textbf{Overview of neurophysiological features and supervised classification performance.} \color{darkgray} The amplitude and latency of peak amplitudes were derived after epoch-wise baseline removal and class-wise averaging of epochs. Values reported for N150 were determined as the minimum of channel O1 of the interval [100, 200]\,ms, while the late positivity (P300) was derived as the maximum of channel Cz in the interval [250 500]\,ms. The last column lists the AUC values estimated via cross-validation from a supervised classifier (see text). } 
\begin{center}
\begin{tabular}{@{\extracolsep{4pt}}@{}cc@{}cc@{}cc}
                                & \multicolumn{2}{c}{\textbf{N150 (O1)}}   & \multicolumn{2}{c}{\textbf{P300 (Cz)}}   &                   \\ \cline{2-3} \cline{4-5} 
\textbf{Subject}   & \textit{Ampl. (µV)} & \textit{Lat. (ms)} & \textit{Ampl. (µV)} & \textit{Lat. (ms)} & \textbf{AUC (\%)}  \\ 
\hline 
S1                       & -9.76               & 150                & 2.72                & 340                & 98.85             \\ 
\rowcolor[HTML]{C0C0C0} 
S2                          & -11.11             & 150                & 1.48                & 400                & 98.73             \\
S3                         & -5.63               & 170                & 1.94                & 500                & 98.06             \\
\rowcolor[HTML]{C0C0C0} 
S4                        & -9.48               & 160                & -0.25               & 500                & 99.82             \\
S5                       & -7.59                & 160                & 1.15                & 410                & 97.05             \\
\rowcolor[HTML]{C0C0C0} 
S6                          & -12.17            & 170                & 0.65                & 470                & 97.12             \\
S7                        & -7.79               & 150                & 1.13                & 450                & 99.92             \\
\rowcolor[HTML]{C0C0C0} 
S8                        & -3.57               & 180                & 3.87                & 360                & 91.69             \\
S9                         & -13.25              & 140                & 0.11                & 380                & 99.56             \\
\rowcolor[HTML]{C0C0C0} 
S10                       & -12.01              & 140                & 3.67                & 380                & 99.72             \\
S11                      & -2.93               & 180                & 1.31                & 300                & 89.18             \\
\rowcolor[HTML]{C0C0C0} 
S12                       & -4.35               & 150                & 3.49                & 370                & 98.89             \\
S13                        & -4.10               & 160                & 3.57                & 370                & 98.45             \\ \hline
\textbf{Mean}       & \textbf{-7.98}      & \textbf{158.46}    & \textbf{1.91}       & \textbf{402.31}    & \textbf{97.46}    \\ \hline
\end{tabular}
\end{center}

\label{tab:neurophysiology}
\end{table}

By training a supervised shrinkage-LDA on this data set and calculating the binary target vs.~non-target classification accuracy based on a 5-fold chronological cross-validation, we obtained an average AUC of 97.5\,\% which indicates a very good SNR of the data set. Note that only the few preprocessing steps mentioned in the preprocessing method section were applied and no artefact removal or adjustment of the classification time intervals was performed.

\subsubsection{Homogeneity} 
\begin{figure}[H]
\begin{center}
\includegraphics[width=\textwidth]{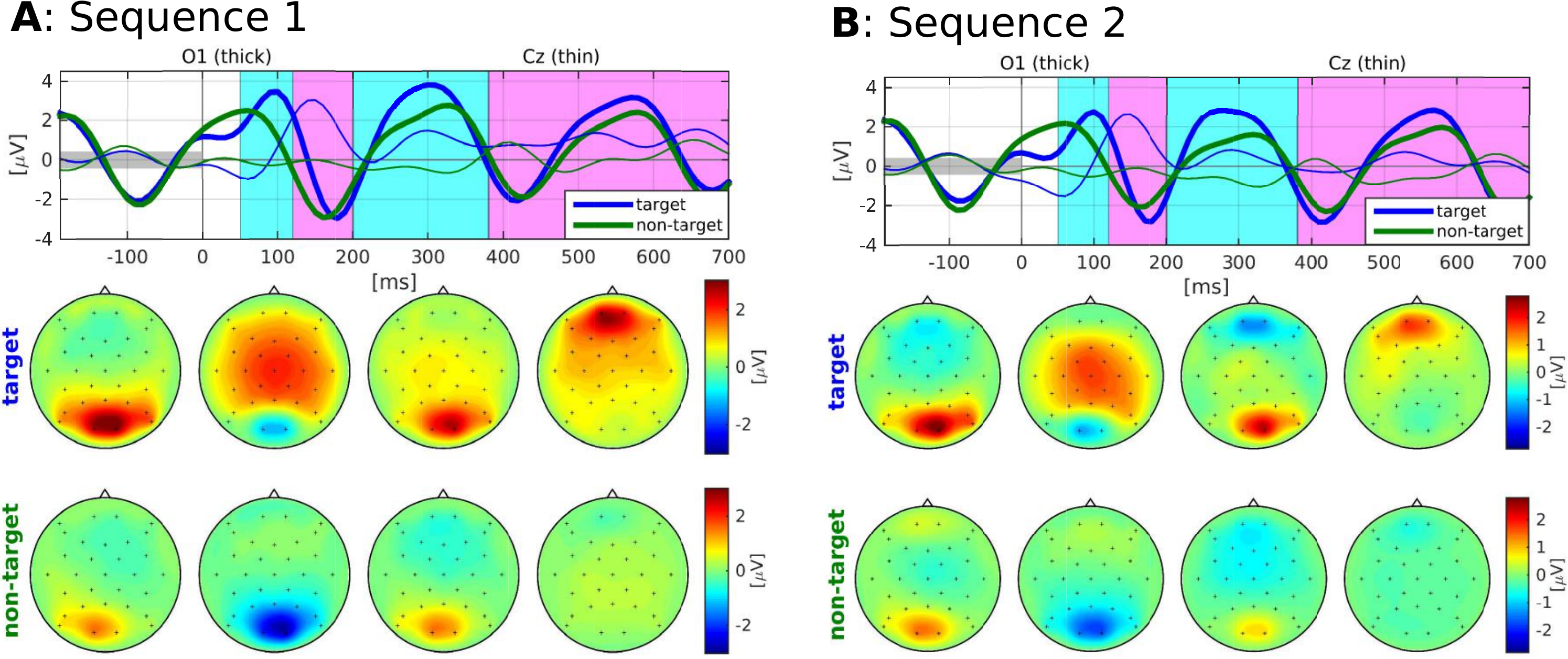}
\end{center}
\caption{\textbf{ERP responses for S11 of sequence 1 \textbf{(A)} and sequence 2 \textbf{(B)}.} \\
\color{darkgray} Top row: Average responses evoked by target (blue) and non-target (green) stimuli in the occipital channel O1 (thick) and the central channel Cz (thin). Prior to averaging, a baseline correction was performed based on data within the interval [-200,0]\,ms. Bottom rows: Scalp plots visualising the spatial distribution of mean target and non-target responses within four selected time intervals: [50 120], [120 200], [201 380] and \mbox{[381 700]}\,ms.}
\label{Fig:ERP_S1_vs_S2}
\end{figure}

To test the homogeneity assumptions of LLP, i.e.~that both sequences have the same average target and non-target ERP responses, we visually inspected the responses for both sequences and each subject with the goal to detect systematic differences in the ERP amplitudes and latencies between the two sequences. Fig~\ref{Fig:ERP_S1_vs_S2} shows the ERP plots from subject S11 for both sequences. Even though small differences can be observed, the ERP responses generally look extremely similar and we could not detect any systematic differences by visual inspection. We also performed a bootstrapping test, as explained in the method section, comparing the similarity of a sample from sequence 1 to the average ERP responses of both sequences. We corrected the significance level by dividing by 13 --- the number of subjects. Note that this is a rather conservative correction as the tests for individual subjects are definitely independent, but the different tests for target and non-target differences of the same subject may be independent as well. We found one significant difference for the corrected significance level ($p^*=0.05/13$), namely the differences in target ERP responses for S4. We will later see that subject S4 nevertheless performed well.

\vspace{-3cm}
\subsubsection{Reconstructed Means} 
\begin{figure}[H]
\begin{center}
\includegraphics[width=\textwidth]{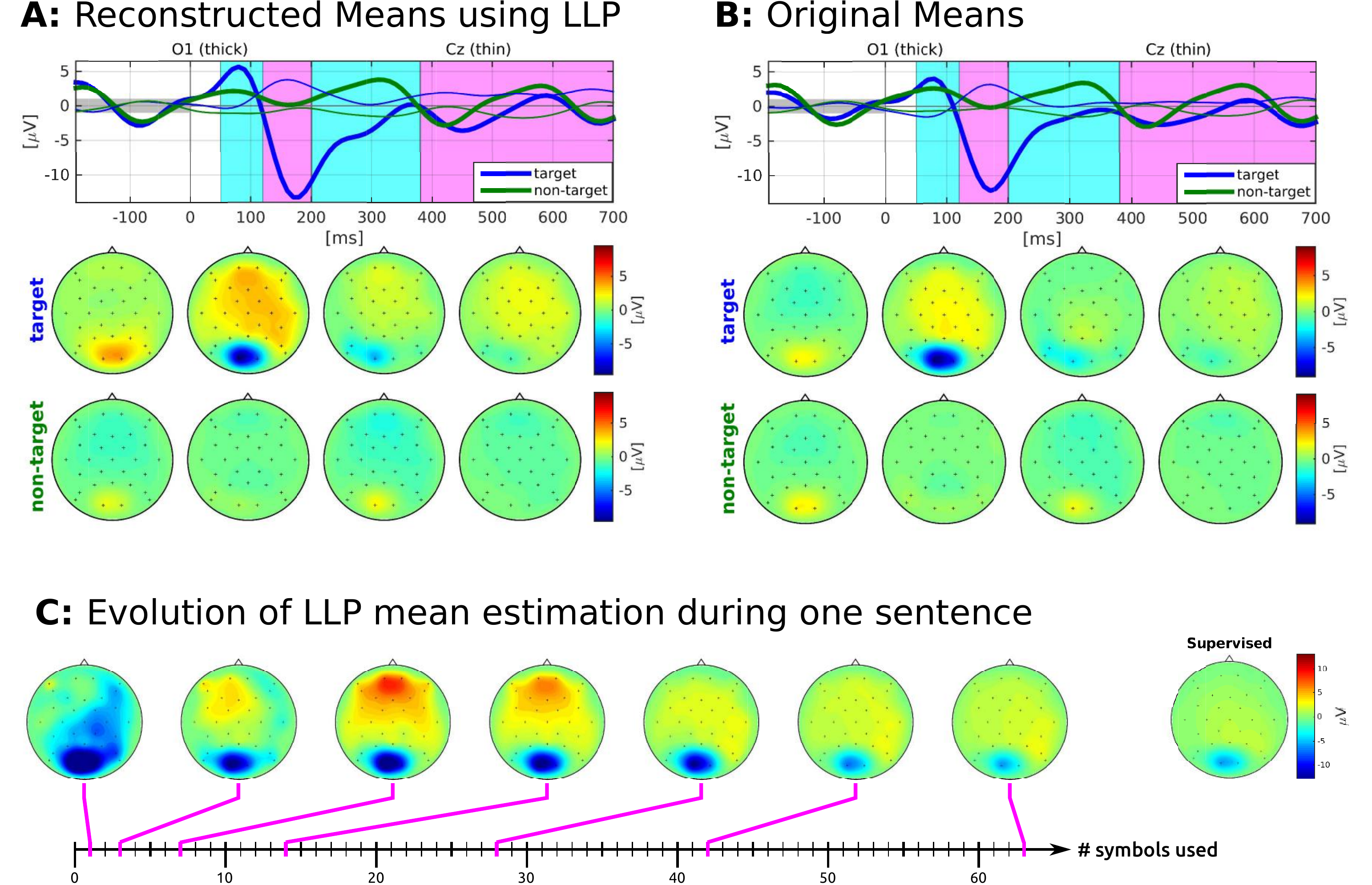}
\end{center}
\caption{\textbf{ERP responses for S6 of the reconstructed class-wise means using LLP (\textbf{A}) and original labelled data (\textbf{B}). C shows the LLP target estimations in [120 200]\,ms for different numbers of training points}. \\ \color{darkgray} For details, see description of Fig~\ref{Fig:ERP_S1_vs_S2}.}
\label{Fig:ERP_orig_vs_reconstruct}
\end{figure}

Next, we investigated if LLP could correctly reconstruct the mean target and non-target ERP responses, when the full amount of data corresponding to three sentences is provided. The ERP plots for subject S6 and four intervals are given in Fig~\ref{Fig:ERP_orig_vs_reconstruct}. It compares the target and non-target ERP means estimated by LLP (Fig~\ref{Fig:ERP_orig_vs_reconstruct}\textbf{A}) with the true class means (Fig.~\ref{Fig:ERP_orig_vs_reconstruct}\textbf{B}). We observe, that the reconstructed class means capture the characteristics of the original means almost flawlessly. 

However, it is also of interest, how the class means estimated by LLP evolve using a growing amount of data. As an example the target mean for subject S6 is provided in Fig.~\ref{Fig:ERP_orig_vs_reconstruct}\textbf{C}. Using epochs that correspond to 1, 3, 7, 14, 28, 42 and 62 symbols, the mean target pattern in the interval [120 200]\,ms stabilizes towards the supervised true mean. While the negative potential over occipital channels undergoes a linear development from strong to weak intensitiy, the activity in frontal and central channels reveals jumps between negative and positive potentials specifically during the first 10 symbols.

\subsubsection{Online Spelling Performance}

\begin{figure}[H]
\centering
\begin{center}
\includegraphics[trim={2.6cm 0 0 0},clip,width=1.1\textwidth]{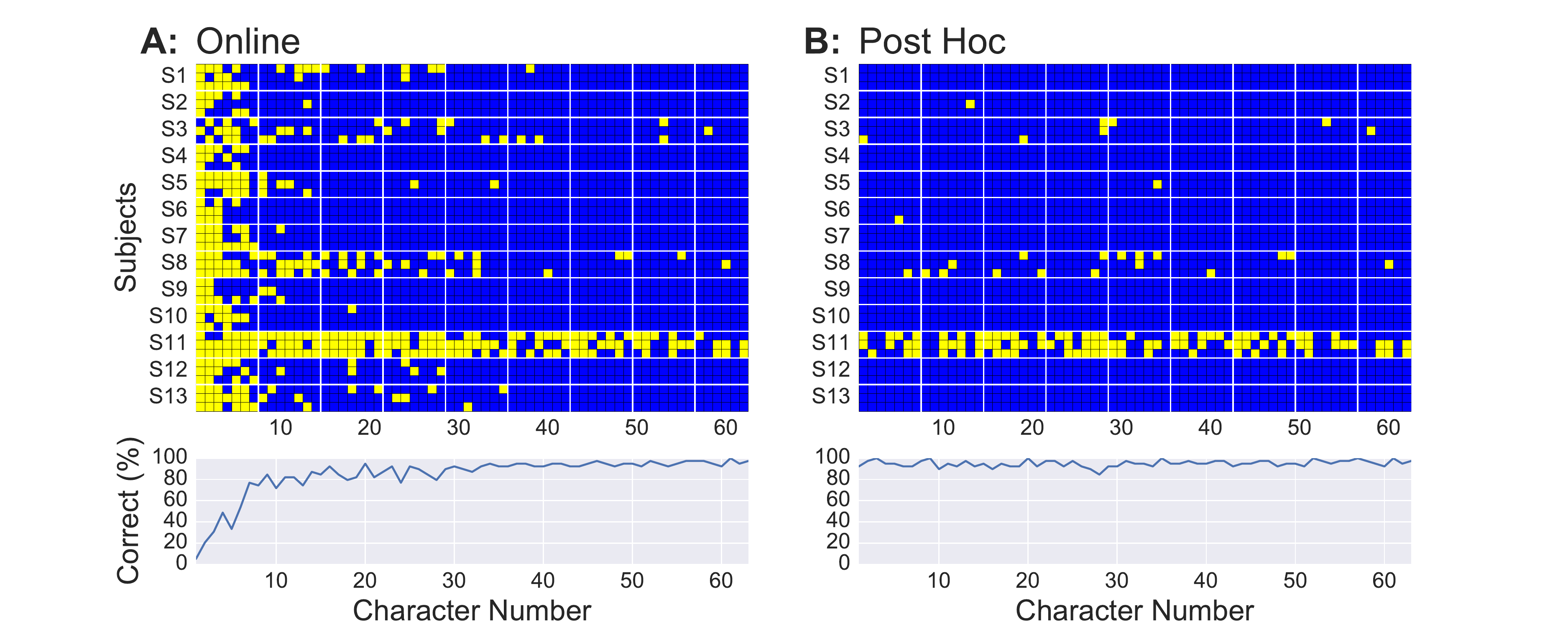}
\end{center}
\caption{{\bf Spelling performance as seen online using the LLP (A) and after the post hoc re-analysis with LLP (B).} \\
\color{darkgray} Top: Each row represents a single spelling of the test sentence "Franzy jagt ...", with yellow squares indicating incorrectly spelled characters and blue squares indicating correctly spelled characters. Bottom: Average number of correctly classified characters.}
\label{Fig:llp_online}
\end{figure}

Fig~\ref{Fig:llp_online}\textbf{A} shows the character-wise online spelling performance with LLP for all 13 subjects including the grand average. In total, 84.5\,\% of all characters were spelled correctly (chance level = 3\,\%). After a ramp-up phase of around 7 characters (which corresponds to 3 minutes wall clock time), this accuracy reaches 90.2\,\% correct characters on the remaining characters on average. In general, the algorithm worked well for all subjects except for S11. The reason for S11's low performance could be determined as an overall low SNR. It is evident also when looking at the supervised performance values provided in Table~\ref{tab:neurophysiology} and by the lack of class-discriminative N150 depicted by Fig~\ref{Fig:ERP_S1_vs_S2}. However, we could not observe that the data of S11 explicitly violated the homogeneity assumption, see also Fig~\ref{Fig:ERP_S1_vs_S2}.

As mentioned before, the advantage of an unsupervised adaptive classifier in a spelling application is that early trials can constantly be re-analysed at any later stage of the spelling, when an improved classifier may be available. A re-evaluation of all characters with the classifier obtained at the end of each sentence is provided by Fig~\ref{Fig:llp_online}\textbf{B}. The post hoc performance of the LLP is extremely high, showing zero or one error for 10 out of 13 subjects. The post hoc classifier is able to resolve the majority of characters misclassified by the online LLP seen in Fig~\ref{Fig:llp_online}\textbf{A}. 

\subsection{Relating LLP to an other Unsupervised Algorithm}
In the previous section, we showed that LLP can successfully be used as a novel classification method in an online study. Given the high SNR of the visual ERP data in the online study, the question remains how well LLP performs in comparison to other unsupervised methods. We chose to compare LLP with another unsupervised algorithm in ERP-BCI which was successfully used in online classification without prior training, namely the unsupervised classification approach based on expectation-maximization (EM) by Kindermans et al.~\cite{kindermans2014true}. The EM algorithm makes use of a probabilistic model which describes the ERP decoding. Although it has no guarantees to converge to a good classifier, it works well in practice when several randomly initialised classifiers are used in parallel. 

The comparison between the LLP- and EM-based unsupervised approaches was done using the data set obtained from our online study. To provide a fair comparison of both classifiers, we simulated an online scenario where both classifiers were retrained after each character and were reset to a random initial state before each sentence. The performance was evaluated on the training set and made use of the label information. Note that over-fitting is less an issue for the two approaches, since no label information are used for training any of the two classifiers. For the EM algorithm, we used the same parameters as described in~\cite{kindermans2014true}.

\begin{figure}[H]
\begin{center}
\includegraphics[width=1\textwidth]{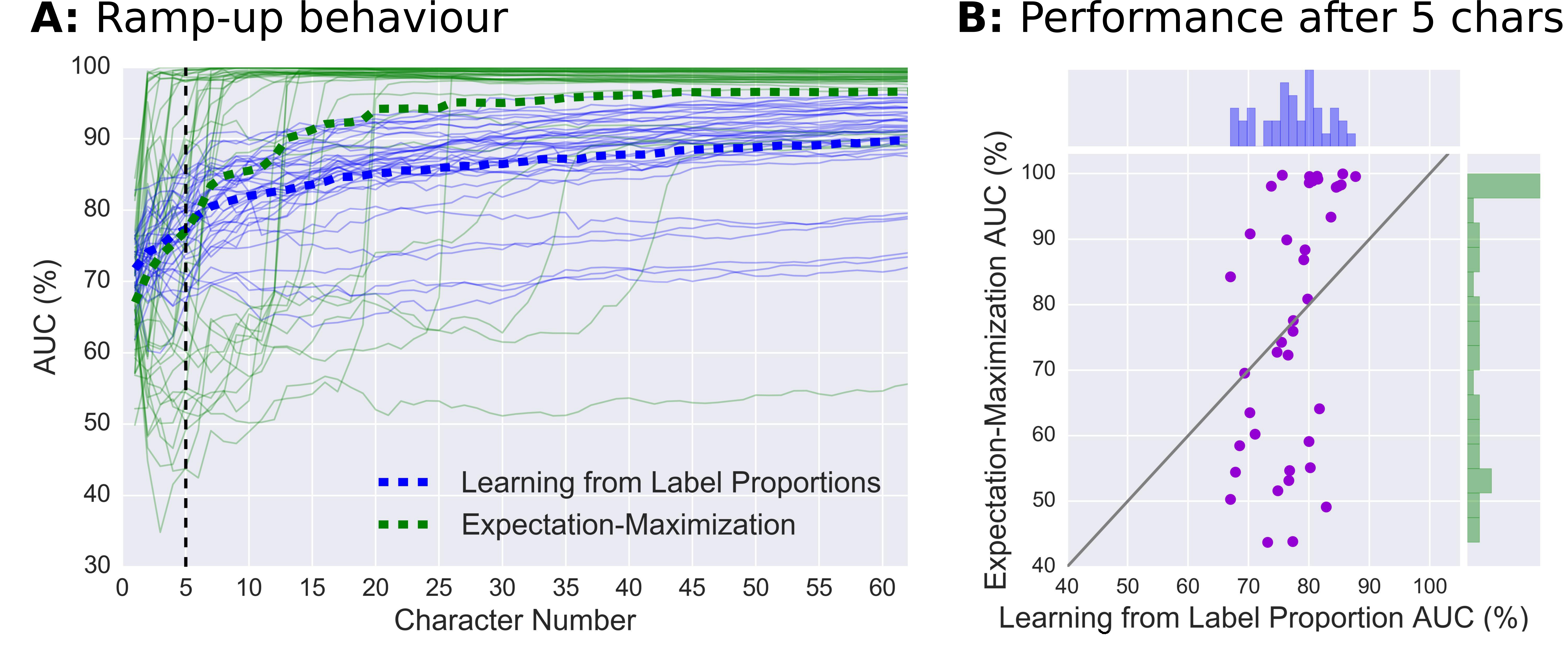}
\end{center}
\caption{{\bf Comparison of LLP to an unsupervised EM-based classification approach for each sentence (A) and after 5 characters (B)} \\ \color{darkgray} \textbf{A}: Thin lines represent the binary target vs. non-target AUC performance of the two learning models with every line corresponding to the spelling of a single sentence. Each of the subjects ($N=13$) spelled each sentence three times resulting in 39 lines. Dashed lines depict average performances. \textbf{B}: Each dot represents the EM and LLP performance after 5 characters for the same subject and sentence.
}
\label{Fig:comparison_LLP_EM}
\end{figure}

The ramp-up performance curves of both classifiers in each of the spelled sentences are depicted in Fig~\ref{Fig:comparison_LLP_EM}\textbf{A}. A comparison of the curve shapes indicates that the two algorithms work in different ways and exploit information contained in the data in different ways. While LLP is constantly improving over time and performs well above chance level for all sentences, the EM algorithm behaves dichotomous: depending on the initialisation, it either works extremely well from an early time point on, or fails to display significant performance increases for a relatively long time period. This dichotomous behaviour can also be seen in Fig~\ref{Fig:comparison_LLP_EM}\textbf{B} where the performance comparison of both classifiers after 5 characters is shown. One can see that the LLP performance for each subject and sentence is between 65\,\% and 90\,\% whereas the EM performance is very spread out with instances below chance level (50\,\%) and cases with almost perfect performance. After having learned on the full data of 63 characters, the EM-based approach outperforms LLP on most sentences.

\vspace{-3cm}
\section{Discussion}
A striking feature of LLP is its low conceptual and computational complexity reducing the probability of implementation errors and run-time problems in an online experiment. The computationally expensive step of generating optimized stimulation sequences can be executed prior to the online experiment.

\subsection{Online Performance of LLP}
Results from the online study showed that the performance of the LLP classifier improves rapidly in the beginning and increases further when more and more data is available. This is accompanied by an increasing quality of the class-wise mean estimations. Since the user can in principle rely on the correct post hoc analysis of initial mistakes, he/she could start spelling without time delay using LLP. This makes it an attractive option in contrast to supervised classifiers which require tedious calibration sessions. One can also observe that at least the initial performance of LLP is competitive to the EM-algorithm by Kindermans et al. As LLP does not rely on random initialisations, it can be perceived as more robust when only a limited amount of data is available. 

\subsection{Mixture Matrix}
Learning from label proportions crucially depends on the possibility to include at least two sequences with different target to non-target ratios into the BCI paradigm. Without this modification, it is not directly applicable to standard ERP paradigms. The best performance and lowest noise amplification factor can be attained, when one sequence predominantly contains targets and the other sequence predominantly contains non-targets. In the limit, this is a supervised scenario. However, it is important to realize that practical limitations come into play when choosing the mixing matrix. For instance, enforcing a target ratio of 1/2 requires a simultaneous highlighting of half the target symbols. If another sequence only consists of non-targets (visual blanks) and the number of highlighted symbols needs to be matched, this would require that half the amount of selectable characters are to be added as visual blanks. This would drastically increase the matrix size. Additionally, if many symbols are highlighted simultaneously, then the number of epochs required to obtain unique decodability of a character increases. Our specific choice of the mixing matrix reflects a trade-off between classifier quality, spelling matrix size and sequence length. However, other (more extreme) choices are possible.

\subsection{Visual Highlighting Scheme}
Comparing to previous studies with visual ERPs, the N150 elicited for target stimuli in this online study is very large~\cite{jin2012combined, townsend2010novel, treder2010visual}, even when using familiar faces as stimuli~\cite{yeom2014efficient, kaufmann2011flashing} or motion onset~\cite{hong2009motiononset}. It may be caused by three factors: First and most importantly, the trichromatic grid overlay is perceived as a very salient stimulus compared to traditional brightness intensifications. While its saliency may have been further enhanced by the short rotation added to the grid, Tangermann et al.~\cite{tangermann2011data} showed, that most of the salience improvement compared to brightness highlighting is caused by the grid effect alone. Second, the SOA of 250\,ms is rather long for a visual paradigm. While target-to-target distance is known as a covariate for P300 amplitude~\cite{gonsalvez2002p300}, longer SOAs may have an affect also to other ERP components. Third, we used precise optical markers to determine stimulus onset time points. Compared to an alternative strategy to use markers elicited by the presentation software, jitter and delay caused by the graphics adapter and the LCD screen are eliminated by the optical markers. This improves the average supervised classification performance by approximately 0.5\,\%.

\subsection{Limitations}
It should be evident that the IID assumption does not hold for features based on EEG signals due to non-stationarities. To counteract them, an adaptive version of the LLP classifier could be implemented similar to the unsupervised adaptation by Vidaurre et al.~\cite{vidaurre2011toward} which gives a higher weight to more recent data points. Alternatively, other techniques to compensate for non-stationarities could be employed such as covariate-shift adaptation~\cite{SugKraMue07} or stationary subspace analysis~\cite{BunMeiKirMue09}. We also observed a violation of the homogeneity assumption for S4, but S4 was nevertheless one of the best subjects.

Assuming the standard Gaussian noise model for ERP features, with Gaussian class distributions and equal covariances~\cite{blankertz2011single}, linear discriminant analysis delivers an optimal classification model with respect to maximizing the probability of correctly assigning a sample to its corresponding class. LLP is limited by the fact that it only reconstructs the class means exactly, but not the class-wise covariance matrices. In our application, we used a heuristic and estimated the class-wise covariance matrices by computing the pooled covariance, and this approximation seemed to work well in practice. Contrary to a linear discriminant analysis, however, the LLP does not fulfil the optimality property of maximizing the probability of correct class predictions.

Another downside of the proposed approach is a reduction of the information transfer as a result of assigning no function to some of the stimuli. This increases the matrix size and the number of highlighted symbols, and reduces the size of each individual symbol on the screen. However, as many factors such as the optimal number of highlighted symbols per stimuli or the optimal matrix size are still unknown, it is hard to quantify this loss. Ultimately, the improved classification performance may compensate for this investment.

\subsection{Possible Extensions}
We see this study only as a proof-of-concept of LLP and believe that extensions based on the LLP principle can be even more valuable. We want to specifically mention two ideas. First, LLP could be combined with other unsupervised methods such as the EM-algorithm. This is based on the observation that both classifiers have their strength during different parts of the experiment and work on different principles. Especially at the start of the ramp-up phase, the probabilistic mean estimation in the EM-algorithm could significantly benefit from the mean estimations obtained from LLP. Following this hypothesis, a combined approach could lead to a faster ramp-up behaviour and a more robust classifier compared to the traditional EM-algorithm or the standalone LLP. This idea was picked up by Verhoeven and colleagues~\cite{verhoeven2017mix} who show that the performance of a combined approach can even transcend the performance of each individual classifier at almost any time. Second, LLP could be used in a transfer learning scenario where one starts with a general classifier obtained on several other subjects and utilizes LLP as an unsupervised adaptation method with guarantees. The extreme simplicity of LLP could facilitate both extensions. Additional extensions such as artefact removal, early stopping or making use of language models can be expected to further increase the spelling speed~\cite{kindermans2014integrating}, but are rather independent of LLP as a classification approach.

\subsection{Application Scenarios}
Going beyond visual ERP paradigms, we briefly want to outline how LLP can also be used for other ERP paradigms such as auditory or haptic ones. LLP works in a general setting with multiple selection options when strictly assigning a non-target function to one of the possible options. For instance, in the 6-class auditory AMUSE paradigm~\cite{schreuder2010new}, one could assign no specific control command to one of the 6 stimuli. Hence, this stimulus would never be attended and always be a non-target. The target proportion for the other stimuli would then be 1 out of 5. This yields two groups with different target to non-target ratios such that LLP can be applied. 

\section{Conclusion}
Experimental paradigms for BCI and machine learning methods usually are developed and applied independently from each other. In our work, we have shown, how an information theoretical requirement of a decoding approach successfully exerts explicit influence onto the experimental protocol of a BCI paradigm, thus optimizing the interaction of the decoding algorithm, user and paradigm as a whole. We have exemplified this strategy by introducing a novel, easy-to-implement, unsupervised learning approach to the BCI community -- learning from label proportions (LLP). Under the assumption of IID data points, the LLP classifier is \textit{guaranteed} to recover class-means of ERP responses, which is not the case for any other unsupervised approach known to the authors.

The experimental protocol of a visual ERP speller was modified by introducing groups of stimulation sequences, with the goal to meet the theoretical requirements of the LLP classifier as good as possible, before we applied it practically in simulations as well as in an online spelling experiment. We found that our protocol adaptation was successful, as the central homogeneity requirement (class-wise means are equal for all stimulation sequences) was violated in only one out of 26 conditions. Even when the violation occurred, the performance was good. Furthermore, we observed that LLP succeeded in estimating the class means from unlabelled data, and found that this classifier works well in practice even though its IID assumption is not realistic for ERP-EEG data.

In the online scenario, 12 out of 13 untrained healthy young participants were able to use the LLP-controlled spelling application without explicit calibration. Comparisons with an EM-based unsupervised classification approach indicate that LLP's performance on small unlabelled data sets is highly competitive, but that its theoretical guarantees come at the cost of slower convergence for larger number of data.

Future work will investigate the combination of LLP with other unsupervised classification and transfer learning approaches. Additionally, application fields outside of visual ERP paradigms will be explored.

\subsection*{Acknowledgements}
DH, KS and MT gratefully acknowledge the support by BrainLinks-BrainTools Cluster of Excellence funded by the German Research Foundation (DFG), grant number EXC 1086. DH and MT further acknowledge the bwHPC initiative, grant INST 39/963-1 FUGG), which provided the computing power for offline data analyses and the support by the Wissenschaftliche Gesellschaft Freiburg im Breisgau. PJK gratefully acknowledges funding from the European Union’s Horizon 2020 research and innovation programme under the Marie Sklodowska-Curie grant agreement NO 657679. TV gratefully acknowledges financial support from the Special Research Fund from Ghent University. KRM gratefully acknowledges funding by the BK21 program funded by Korean National Research Foundation grant (No. 2012-005741).

\end{document}